\documentclass[twocolumn]{article} 
\usepackage{graphicx}
\usepackage{amsmath}
\usepackage{amssymb}
\usepackage{url}
\usepackage{balance}
\usepackage{ctable}
\usepackage{multirow}
\usepackage{stfloats}
\usepackage{array}
\usepackage{natbib}
\newcolumntype{Q}{>{\centering\arraybackslash}m{1cm}}
\newcolumntype{O}{>{\centering\arraybackslash}m{0.8cm}}
\newcolumntype{V}{>{\centering\arraybackslash}m{1.62cm}}
\newcolumntype{R}{>{\centering\arraybackslash}m{2.2cm}}
\newcolumntype{L}{>{\centering\arraybackslash}m{3cm}}
\newcolumntype{S}{>{\raggedright\arraybackslash}m{3.8cm}}
\begin{document}

\title{Real Time Surveillance for Low Resolution and Limited-Data Scenarios: An Image Set Classification Approach}

\author{Uzair Nadeem\thanks{Department of Computer Science and Software Engineering, The University of Western Australia. uzair.nadeem@research.uwa.edu.au}  \and 
Syed Afaq Ali Shah\thanks{Department of Computer Science and Software Engineering, The University of Western Australia. afaq.shah@uwa.edu.au}  \and 
Mohammed Bennamoun\thanks{Department of Computer Science and Software Engineering, The University of Western Australia. mohammed.bennamoun@uwa.edu.au}  \and
Roberto Togneri\thanks{Department of Electrical, Electronics and Computer Engineering, The University of Western Australia. roberto.togneri@uwa.edu.au}  \and 
Ferdous Sohel\thanks{School of Engineering and Information Technology, Murdoch University. f.sohel@murdoch.edu.au}
}

\date{}

\maketitle

\begin{abstract}
This paper proposes a novel image set classification technique based on the concept of linear regression. Unlike most other approaches, the proposed technique does not involve any training or feature extraction. The gallery image sets are represented as subspaces in a high dimensional space. Class specific gallery subspaces are used to estimate regression models for each image of the test image set. Images of the test set are then projected on the gallery subspaces. Residuals, calculated using the Euclidean distance between the original and the projected test images, are used as the distance metric. Three different strategies are devised to decide on the final class of the test image set. We performed extensive evaluations of the proposed technique under the challenges of low resolution, noise and less gallery data for the tasks of surveillance, video-based face recognition and object recognition. Experiments show that the proposed technique achieves a better classification accuracy and a faster execution time compared to existing techniques especially under the challenging conditions of low resolution and small gallery and test data.
\end{abstract}

\section*{Keywords}
Image Set Classification $\quad$ Linear Regression Classification$\quad$ Surveillance$\quad$ Video-based Face Recognition$\quad$ Object Recognition

\section{Introduction}\label{sec:introduction}

In the recent years, due to the wide availability of mobile and handy cameras, image set classification has emerged as a promising research area in computer vision. The problem of image set classification is defined as the task of recognition from multiple images \citep{kim2007discriminative}. Unlike single image based recognition, the test set in image set classification consists of a number of images belonging to the same class. The task is to assign a collective label to all the images in the test set. Similarly, the training or gallery data is also composed of one or more image sets for each class and each image set consists of different images of the same class.

Image set classification can provide solution to many of the shortcomings of single image based recognition methods. The extra information (poses and views) contained in an image set can be used to overcome a wide range of appearance variations such as non-rigid deformations, viewpoint changes, occlusions, different background and illumination variations. It can also help to take better decisions in the presence of noise, distortions and blur in images.
These attributes render image set classification as a strong candidate for many real life applications e.g., surveillance, video-based face recognition, object recognition and person re-identification in a network of security cameras \citep{hayat2015deep}.

The image set classification techniques proposed in the literature can broadly be classified into three categories.  \textbf{(i)} The first category, known as parametric methods, uses statistical distributions to model an image set \citep{arandjelovic2005face}. These techniques calculate the closeness between the distributions of training and test image sets. These techniques however, fail to produce good results in the absence of a strong statistical relationship between the training and the test data. \textbf{(ii)} Contrary to parametric methods, non-parametric techniques do not model image sets with statistical distributions. Rather, they represent image sets as linear or nonlinear subspaces \citep{wang2012covariance,ortiz2013face,yang2013face,zhu2013point}. \textbf{(iii)} The third type of techniques use machine learning to classify image sets \citep{hayat2014learning,hayat2015deep,lu2015multi}.

Most current image set classification techniques have been evaluated on video-based face recognition. However, the results cannot be generalized to real life surveillance scenarios, because the images obtained by surveillance cameras are usually of low resolution and contrast, and contain blur and noise. Moreover, in surveillance cases, backgrounds in training and testing are usually different. The training set usually consists of high resolution mug shots in indoor studio environments, while the test set contains noisy images in highly uncontrolled environments. Another challenge is that not enough data is normally available for testing and training, e.g., even if there is a video of a person to be identified, captured by a surveillance camera, there may be very few frames in which the face of that person is recognizable. In addition to humans, it may also be required to recognize various objects in a surveillance scenario.

The current state-of-the-art techniques for image set classification are learning-based and they require a lot of training data to achieve good results. However, in practical scenarios (e.g., to identify a person captured in a surveillance video from a few mugshots) very little training data is usually available per class. Moreover, the current techniques are not able to cope effectively with the challenges, e.g., of low resolution images. Image set classification techniques usually require hand-crafted features and parameter tuning to achieve good results. Computational time is another constraint as decisions for surveillance and security are required in real-time. Moreover, it is difficult to add new classes in methods that require training, as this will require the re-training on all of the data. An ideal image set classification technique should be able to distinguish image sets with less gallery data and should be capable of coping with the challenges of low resolution, noise and blur. The technique should also work in real time with minimal or no training (or parameter tuning) with the capability to easily add new classes.

\begin{figure*}[t]
\begin{center}
   \includegraphics[width=1\linewidth]{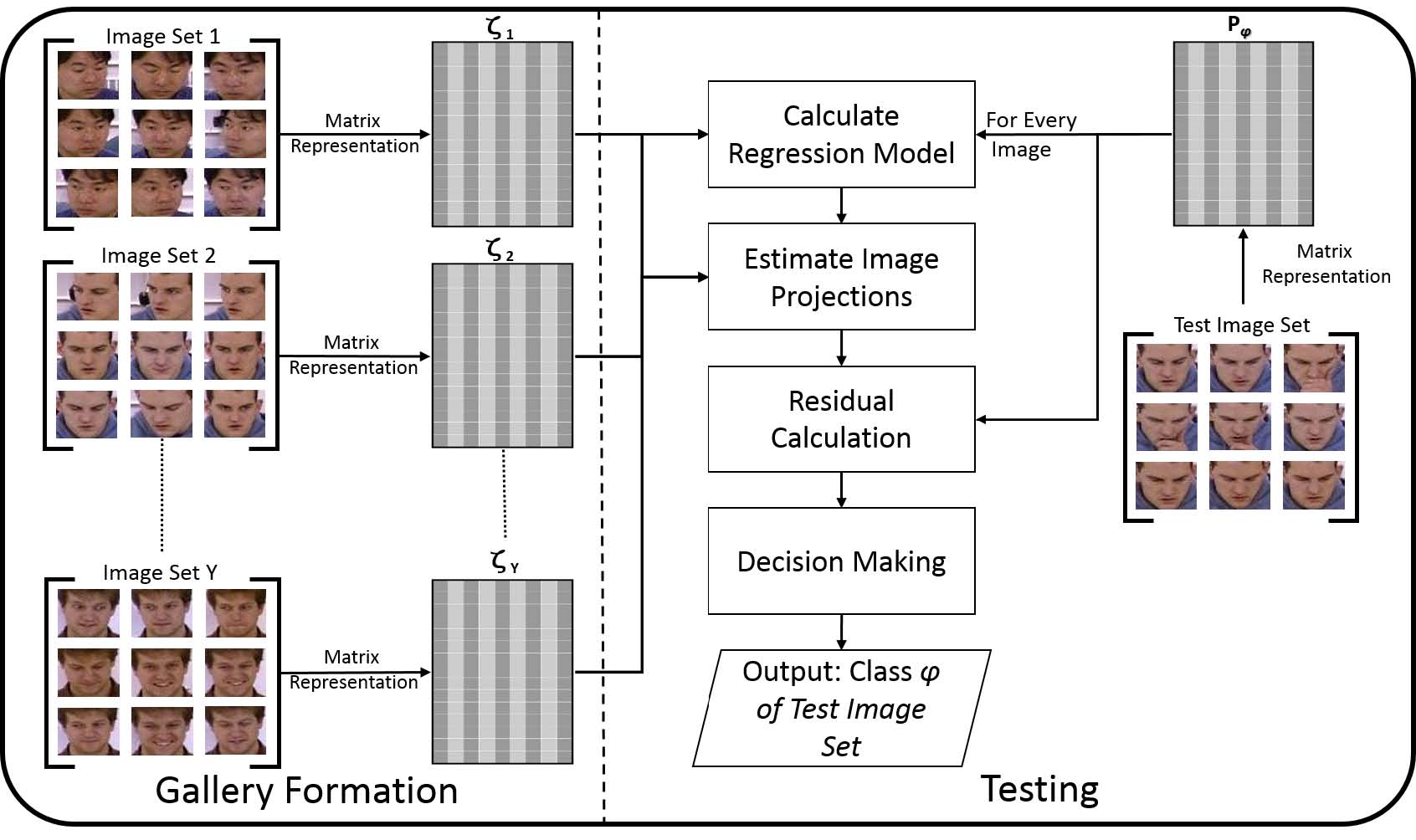}
\end{center}
   \caption{A block diagram of the proposed technique. Each gallery image set is converted to a matrix and is treated as a subspace. Each test image is projected on the gallery subspaces. The residuals, obtained by the difference of the original and the projected images, are used to decide on the class of the test image set.}
\label{fig:blockDig}
\end{figure*}

This paper proposes a novel non-parametric approach for image set classification. The proposed technique is based on the concept of image projections on subspaces using Linear Regression Classification (LRC) \citep{naseem2010linear}. LRC uses the principle that images of the same class lie on a linear subspace \citep{belhumeur1997eigenfaces,basri2003lambertian}. In our proposed technique, we used downsampled, raw grayscale images of gallery set of each category to form a subspace in a high dimensional space. Our technique does not involve any training. At test time, each test image is represented as a linear combination of images in each gallery image set. Regression model parameters are estimated for each image of the test image set using a least squares based solution. Due to the least squares solution, the proposed technique can be susceptible to the problem of singular matrix or singularity. This occurs when the matrix is rank deficient i.e., the rank of the matrix is less than the number of rows. We used perturbation to overcome this problem. The estimated regression model is used to calculate the projection of the test image on the gallery subspace. The residual obtained for the test image projection is used as the distance metric. Three different strategies: majority voting, nearest neighbour classification and weighted voting, are evaluated to decide for the class of the test image set. The block diagram of the proposed technique is shown in Figure \ref{fig:blockDig}.
We carried out a rigorous performance evaluation of the proposed technique for the tasks of real life surveillance scenarios on the SCFace Dataset \citep{scfacegrgic2011}, FR\_SURV Dataset \citep{frsurv} and Choke Point Dataset\citep{choke}. We also evaluated our algorithm for the task of video-based faced recognition on three popular image set classification datasets: CMU Motion of Body (CMU MoBo) Dataset \citep{gross2001cmu}, YoutubeCelebrity (YTC) Dataset \citep{kim2008face} and UCSD/Honda Dataset \citep{lee2003video}. We used the Washington RGBD Dataset \citep{wrgbd} and  ETH-80 dataset \citep{leibe2003analyzing} for experiments on object recognition. The experiments are carried out at different resolution and for different sizes of gallery and test sets to validate the performance of the proposed technique for low resolution and limited data scenarios. We also provide comparison with other prominent image set classification algorithms.

The main contribution of this paper is a novel image set classification technique that is able to produce a superior classification performance, especially under the constraints of low resolution and limited gallery and test data. Our technique does not require any training and can easily be generalized across different datasets. A preliminary version of this work for video-based face recognition appeared in \cite{mypaper}. However, the experiments in \cite{mypaper} were mainly geared towards video-based face recognition datasets at normal resolutions and did not evaluate the ability of the proposed technique for real life surveillance scenarios e.g., situations in which the gallery and the test data have been acquired under significantly different conditions such as, camera noise and scale (i.e., distance of the imaging device from the object). This paper extends \cite{mypaper} by improving the decision making process and provides three different strategies for the estimation of the class of the test image set. This paper also provides a more detailed explanation of the proposed technique and adds extensive experiments to demonstrate the superior capability of the proposed technique to perform effectively using a limited number of low resolution images. The experiments are performed at three different resolutions to test the performance of our technique under the changes in resolution. To the best of our knowledge, this is the first implementation of image set classification for real life surveillance scenarios. We also evaluated the proposed technique for the tasks of set-based object classification. The effect of variations in the resolution and the size of the gallery, and the test data, on the accuracy and the computational efficiency of the proposed technique is also presented, along with the comparison with other techniques. The performance of the proposed technique on three different types of datasets (surveillance, video-based face recognition and set-based object recognition) also establishes the generalization capabilities of the technique. Also in \cite{mypaper}, the experimental results of the compared techniques were extracted from \cite{hayat2015deep}. However, due to variations in the testing protocols and in the data used for experiments, there are significant differences in the results. This issue is also mentioned by \cite{chen2014dual}. Due to this, it has become a common practice in image set classification works to directly compare with the authors' implementation of the compared methods (e.g., in \cite{yang2013face,zhu2013point,hayat2014learning,hayat2015deep,lu2015multi}) to present a fair comparison. In this paper we have adopted the same approach and directly compared our technique against the original implementations as is provided by the respective authors.

The rest of this paper is organized as follows. Section \ref{related_work} presents an overview of the related work. The proposed technique is discussed in Section \ref{technique}. Section \ref{experiments} reports the experimental conditions and a detailed evaluation of our results, compared to the state-of-the-art approaches. The comparison of the computational time of the proposed technique with other methods is presented in the Section \ref{timing_analysis}. The paper is concluded in Section \ref{conclusion}.

\section{Related Work}\label{related_work}
Image set classification methods can be classified into three broad categories: parametric, non-parametric and machine learning based methods, depending on how they represent the image sets. The parametric methods represent image sets with a statistical distribution model \citep{arandjelovic2005face} and then use different approaches, such as KL-divergence, to calculate the similarity and differences between the estimated distribution models of gallery and test image sets. Such methods have the drawback that they assume a strong statistical relationship between the image sets of the same class and a weak statistical connection between image sets of different classes. However, this assumption does not hold many a times. 

On the other hand, non-parametric methods model image sets by exemplar images or as geometric surfaces. Different works have used a wide variety of metrics to calculate the set to set similarity. \cite{wang2008manifold} use the Euclidean distance between the mean images of sets as the similarity metric. \cite{cevikalp2010face} adaptively learn samples from the image sets and use affine hull or convex hull to model them. The distance between image sets is called Convex Hull Image Set Distance (CHISD) in the case of convex hull model, while in the case of affine hull model, it is termed as the Affine Hull Image Set Distance (AHISD). \cite{hu2012face} calculate the mean image of the image set and affine hull model to estimate the Sparse Approximated Nearest Points (SANP). The calculated Nearest Points determine the distance between the training and the test image sets.  
\cite{chen2012dictionary} estimate different poses in the training set and then iteratively train separate sparse models for them. \cite{chen2013video} use a non-linear kernel to improve the performance of \cite{chen2012dictionary}. The non-availability of all the poses in all videos hampers the final classification accuracy.
Some non-parametric methods e.g., \citep{wang2008manifold,wang2009manifold,harandi2011graph,wang2012covariance}, represent the whole image set with only one point on a geometric surface. 
The image set can also be represented either by a combination of subspaces or on a complex non-linear manifold. In the case of subspace representation, the similarity metric usually depends on the smallest angles between any vector in one subspace and any other vector in the other subspace. Then the sum of the cosines of the smallest angles is used as the similarity measure between the two image sets.  
In the case of manifold representation of image sets, different metrics are used for distance, e.g., Geodesic distance \citep{turaga2011statistical,hayat2014automatic}, the projection kernel metric on the Grassman manifold \citep{hamm2008grassmann}, and the log-map distance metric on the Lie group of Riemannian manifold \citep{harandi2012kernel}.
Different learning techniques based on discriminant analysis are commonly used to compare image sets on the manifold surface such as Discriminative Canonical Correlations (DCC) \citep{kim2007discriminative}, Manifold Discriminant Analysis (MDA) \citep{wang2009manifold}, Graph Embedding Discriminant Analysis (GEDA) \citep{harandi2011graph} and Covariance Discriminative Learning (CDL) \citep{wang2012covariance}. Chen \citep{chen2014dual} assumes a virtual image in a high dimensional space and used the distance of the training and test image sets from the virtual image as a metric for classification. \cite{feng2016pairwise} extends the work of \cite{chen2014dual} by using the the distance between the test set and unrelated training sets, in addition to the distance between the test image set and the related training set. However, these methods can only work for very small image sets due to the limitation that the dimension of the feature vectors should be much larger than the sum of the number of images in the gallery and the test sets.

\cite{gcr} use the relationship between image sets to model the inter-set and intra-set variations. They learn Group Collaborative Representations (GCRs) using a self-representation learning method, to model the probe sets using all the gallery data. \cite{bsvm} use a subset of gallery images and query set to train a binary SVM to learn a representation of the query set. This representation is then used to determine the similarity between the query and the gallery sets. However, their method is computationally expensive at the test time and it is not suitable for small query or gallery sets, since SVM is not able to learn a good representation in such a case. \cite{lu2015multi} use deep learning to learn non-linear models and then apply discriminative and class specific information for classification. \cite{hayat2014learning,hayat2015deep} propose an Adaptive Deep Network Template (ADNT) which is a deep learning based method. They use autoencoders to learn class specific non-linear models of the training sets. Instead of random initialization, Gaussian Restricted Boltzmann Machine (GRBM) is used to initialize the weights of the autoencoders. At test time, the class specific autoencoder models are used to reconstruct each image of the test set, and then the reconstruction error is used as a classification metric. ADNT has been shown to achieve good classification accuracies, but it requires the fine-tuning of several parameters and the extraction of hand crafted LBP features to achieve a good performance. Learning-based techniques are computationally expensive and require a large amount of training data to produce good results.

Our proposed technique estimates the linear regression model parameters for each image in the test set and then reconstructs test images from the gallery subspaces. Our technique can cope well with the challenges of low resolution, noise and small gallery and test data, and does not have any constraints on the number of images in the test set. Moreover, it does not involve any feature extraction or training and can produce state of the art results using only the raw images. At the same time our technique is computationally efficient and can classify image sets much faster than the learning based techniques both at the training and the test times (Section \ref{timing_analysis}). It can also effectively utilize parallel processing or GPU computation to further reduce the computational time.

\section{Proposed Technique}\label{technique}
In this paper, we adopt the following notation: bold capital letters denote $\textbf{MATRICES}$, bold lower letters show $\textbf{vectors}$, while normal letters are for $scalar\ numbers$. We use $[i]$ to denote the set of integer numbers from $1$ to $i$, for a given integer $i$.

Consider the problem to classify the unknown class of a test image set as one of the $[Y]$ classes.
Let the gallery image set of each unique class $[Y]$ contain $N_{\zeta}$ images.
\begin{figure}[t]
\begin{center}
   \includegraphics[width=1\linewidth]{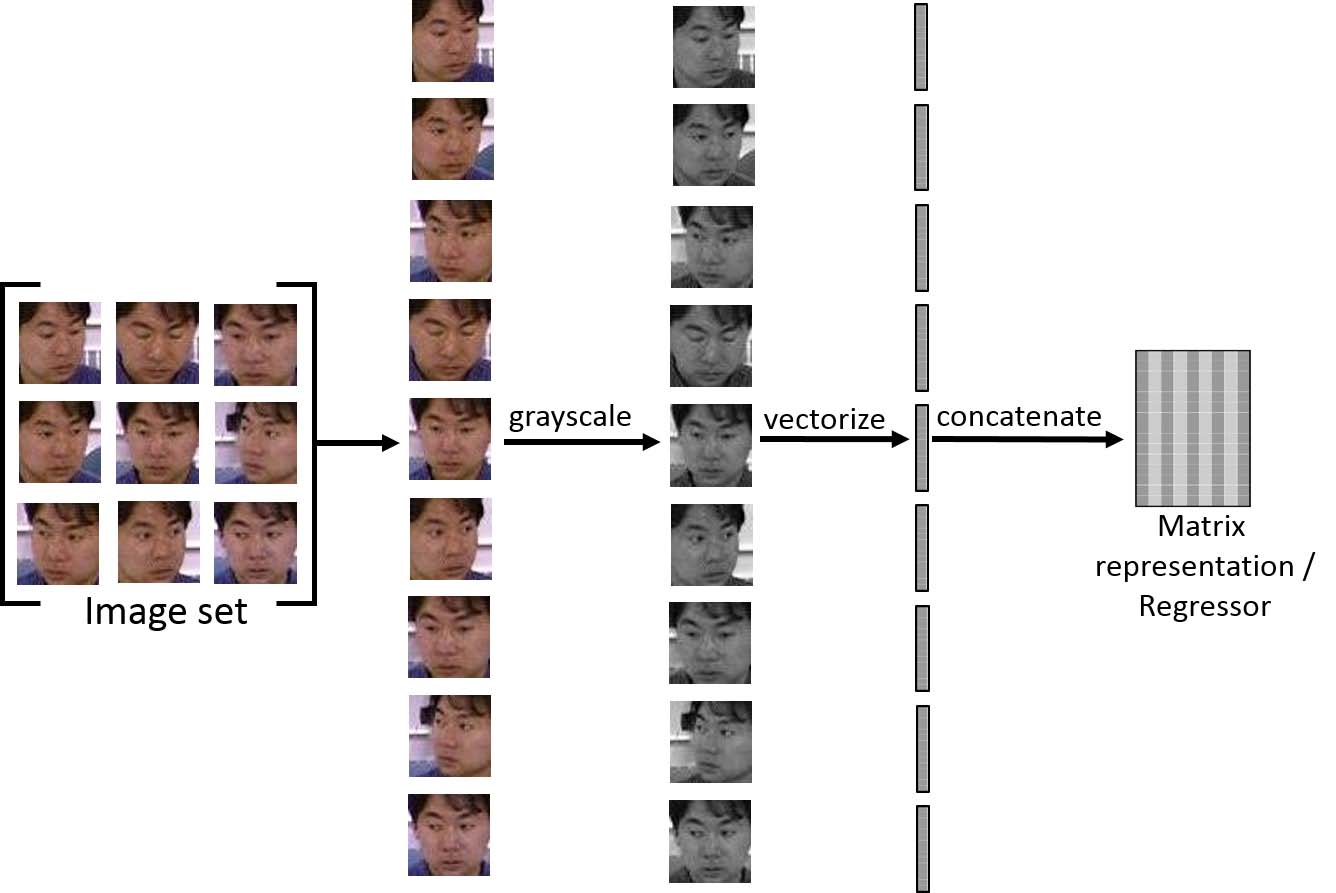}
\end{center}
   \caption{A representation of the regressor formation in the proposed technique.}
\label{fig:MatrixFormation}
\end{figure}
Each image in the gallery image sets is downsampled from the initial resolution of $a \times b$ to the resolution of $c \times d$ and are converted to grayscale to be represented as $\boldsymbol{G_{y}^{i}} \in \mathbb{R}^{c\times d}$, $\forall\ y\in [Y]$ and $\forall\ i\in [N_{\zeta}]$. 
We concatenate all the columns of each image to obtain a feature vector such that $\boldsymbol{G_{y}^{i}} \in \mathbb{R}^{c\times d} \rightarrow \boldsymbol{\varsigma_{y}^{i}} \in \mathbb{R}^{\tau\times 1}$, where $\tau = cd$. We do not use any feature extraction technique and use raw grayscale pixels as the features. Next, an image set matrix $\boldsymbol{\zeta_{y}}$ is formed for each class in $[Y]$ by horizontally concatenating the image vectors of class $y$. 
\begin{equation}
\label{TrainMat}
\boldsymbol{\zeta_y} = [\boldsymbol{\varsigma_y^1 \varsigma_y^2 \varsigma_y^3} ... \boldsymbol{\varsigma_y^{N_{\zeta}}}]\in \mathbb{R}^{\tau\times N_{\zeta}},\quad \forall\ y\in [Y] 
\end{equation}
This follows the principle that patterns from the same class form a linear subspace \citep{basri2003lambertian}. 
Each vector $\boldsymbol{\varsigma_y^i},\ \forall\ i\in [N_{\zeta}],\:$ of the matrix $\boldsymbol{\zeta_y}$, spans a subspace of $\mathbb{R}^{\tau \times 1}$. Thus, each class in $[Y]$ is represented by  a vector subspace $\boldsymbol{\zeta_y}$ called the \textit{regressor} for class $y$. A representation of the regressor formation is shown in
\ref{fig:MatrixFormation}.

Let $\varphi$ be the unknown class  of the test image set with $N_{P}$ number of images.
In the same way as the gallery images, images of the test image set are downsampled to the resolution of $c \times d$ and converted to grayscale to be represented as $\boldsymbol{T_{\varphi}^{j}} \in \mathbb{R}^{c\times d}$ where ${\varphi}$ is the unknown class of the test image set and $j\in [N_{P}]$.
Each downsampled image is converted to a vector through column concatenation such that $\boldsymbol{T_{\varphi}^{j}} \in \mathbb{R}^{c\times d} \rightarrow \boldsymbol{\rho_{\varphi}^{j}} \in \mathbb{R}^{\tau\times 1}, $ where $\tau = cd$. A matrix $\boldsymbol{P_{\varphi}}$ is formed for the test image set by the horizontal concatenation of image vectors $\boldsymbol{\rho_{\varphi}^{j}},\forall\ j\in [N_{P}]$.
\begin{equation}
\label{TestMat} 
\boldsymbol{P_{\varphi}} = [\boldsymbol{\rho_{\varphi}^1 \rho_{\varphi}^2 \rho_{\varphi}^3} ... \boldsymbol{\rho_{\varphi}^{N_{P}}}]\in \mathbb{R}^{\tau\times N_{P}}
\end{equation}
If $\boldsymbol{P_{\varphi}}$ belongs to the $y^{th}$ class, then its vectors should span the same subspace as the image vectors of the gallery set $\boldsymbol{\zeta_y}$. In other words, image vectors of test set $\boldsymbol{P_{\varphi}}$ should be represented as a linear combination of the gallery image vectors from the same class. Let $\boldsymbol{\theta_y^j} \in \mathbb{R}^{N_{\zeta}\times 1}$ be a vector of parameters. Then the image vectors of test set can be represented as:
\begin{equation}
\label{EqtoSolv} 
\boldsymbol{\rho_{\varphi}^j}=\boldsymbol{\zeta_y}\boldsymbol{\theta_y^j},\quad \forall\ j\in [N_{P}],\ \forall\ y\in [Y]
\end{equation}  
The number of features must be greater than or equal to the number of images in each gallery set i.e., $\tau\geq N_{\zeta}$ in order to calculate a unique solution for Equation (\ref{EqtoSolv}) by the Least Squares method \citep{friedman2001elements,ryan2008modern,seber2012linear}. 
However, even if the condition of $\tau\geq N_{\zeta}$ holds, it is possible for one or more rows of the regressor $\boldsymbol{\zeta_y}$ to be linearly dependent on one another, which renders the regressor matrix $\boldsymbol{\zeta_y}$ to be singular. In this case, the regressor $\boldsymbol{\zeta_y}$ is called a rank deficient matrix since the rank of $\boldsymbol{\zeta_y}$ is less than the number of rows. In this scenario, it is not possible to use the least squares solution to estimate a unique parameter vector $\boldsymbol{\theta_y^j}$. The singularity of the regressor $\boldsymbol{\zeta_y}$ can be removed by perturbation \citep{wang2012covariance}. In the case of a singular matrix, a small perturbation matrix $\boldsymbol{\varepsilon} $ with uniform random values in the range $-0.5\leq \epsilon \leq +0.5$ was added to the regressor $\boldsymbol{\zeta_y}$:
\begin{equation}
\label{perturbation}
\boldsymbol{\widetilde{\zeta_y}}=\boldsymbol{\zeta_y}+\boldsymbol{\varepsilon}, \quad \ \boldsymbol{\varepsilon} \in \mathbb{R}^{\tau\times N_{\zeta}} ,\quad \forall \epsilon \in \boldsymbol{\varepsilon} :\  \vert\epsilon\vert \leq 0.5
\end{equation}

\begin{table*}[htbp]
\caption{Average percentage classification accuracies for the task of surveillance on SCface \citep{scfacegrgic2011}, FR\_SURV \citep{frsurv} and Choke Point \citep{choke} datasets.}
  \centering
      \begin{tabular}{|S|Q|Q|Q|Q|Q|Q|Q|Q|Q|}
    \hline
    \multicolumn{1}{|c|}{\textbf{Datasets$\rightarrow$}} & \multicolumn{3}{c|}{\textbf{SC\_Face}} & \multicolumn{3}{c|}{\textbf{FR\_SURV}} & \multicolumn{3}{c|}{\textbf{Choke Point}} \\
    \hline
    \multicolumn{1}{|l|}{\textbf{Methods$\downarrow$ \textbackslash $\ $Resolutions$\rightarrow$}} & \textbf{10$\boldsymbol{\times}$10} & \textbf{15$\boldsymbol{\times}$15} & \textbf{20$\boldsymbol{\times}$20} & \textbf{10$\boldsymbol{\times}$10} & \textbf{15$\boldsymbol{\times}$15} & \textbf{20$\boldsymbol{\times}$20} & \textbf{10$\boldsymbol{\times}$10} & \textbf{15$\boldsymbol{\times}$15} & \textbf{20$\boldsymbol{\times}$20} \\
    \hline
    \textbf{AHISD \citep{cevikalp2010face}} & 30.76 & 40.76 & 40    & 31.37 & 33.33 & 35.29 & 52.27 & 54.54 & 55.08 \\
    \hline
    \textbf{RNP\citep{yang2013face}} & 29.23 & 36.92 & 36.15 & 27.45 & 33.33 & 35.29 & 46.84 & 48.82 & 49.27 \\
    \hline
    \textbf{SSDML \citep{zhu2013point}} & 9.23  & 15.38 & 13.85 & 11.76 & 7.84  & 7.84  & 53.82 & 47.73 & 48.34 \\
    \hline
    \textbf{DLRC\citep{chen2014dual}} & 36.92 & 39.23 & 38.46 & 5.88  & 9.8   & 3.92  & 41.79  & 45.35 & 45.40 \\
    \hline
    \textbf{DRM \citep{hayat2015deep}} & 2.31  & 6.15  & 4.62  & 5.88  & 9.8   & 9.8   & 32.54 & 50.00 & 55.53 \\
    \hline
    \textbf{PLRC \citep{feng2016pairwise}} & 24.62 & 25.38 & 26.15 & 37.25 & 35.29 & 35.29 & 53.85 & 56.19 & 56.83 \\
    \hline
     \textbf{PDL \citep{pdl}} & 8.46 & 9.23 & 9.23 & 23.53 & 27.45 & 27.45 & 38.32 & 41.31 & 42.70 \\
    \hline
     \textbf{DARG \citep{darg}} & 0.77 & 2.31 & 1.54 & 9.80 & 9.80 & 11.76 & 22.75 & 30.11 & 30.19 \\
    \hline
    \textbf{Ours (MV)} & 41.54 & 46.92 & 42.31 & \textbf{45.10}  & 49.02   & 45.10  & 55.11  & 54.95    & 56.34 \\
    \hline
    \textbf{Ours (NN)} & 41.54 & 46.92 & 42.31 & 39.22  & 49.02   & \textbf{47.06}  & 55.99  & 56.68    & 56.98 \\
    \hline
    \textbf{Ours (EWV)} & \textbf{44.62} & \textbf{48.46} & \textbf{46.15} & 41.18 & \textbf{51} & \textbf{47.06} & \textbf{56.42} & \textbf{56.95} & \textbf{57.22} \\
    \hline
    \end{tabular}%
    \vspace{0.2cm}
    
  \label{table:surveillance}%
\end{table*}%
The regressor $\boldsymbol{\zeta_y}$ is perturbed while its values are still in the range of $0$ to $255$. Therefore, the maximum possible change in the value of any pixel is $0.5$. 
Now, the vector of parameters $\boldsymbol{\theta_y^j}$ can be estimated for the test image vector $\boldsymbol{\rho_{\varphi}^j}$ and regressor $\boldsymbol{\widetilde{\zeta_y}}$ by the following equation:
\begin{equation}
\label{Solv1} 
\boldsymbol{\theta_y^j} =(\boldsymbol{\widetilde{\zeta_y}}'\boldsymbol{\widetilde{\zeta_y}})^{-1}\boldsymbol{\widetilde{\zeta_y}}'\boldsymbol{\rho_{\varphi}^j},\quad \forall\ j\in [N_{P}],\ \forall\ y\in [Y]
\end{equation}  
where $\boldsymbol{\widetilde{\zeta_y}}'$ is the transpose of $\boldsymbol{\widetilde{\zeta_y}}$. Next, the estimated vector of parameters $\boldsymbol{\theta_y^j}$ and the regressor $\boldsymbol{\widetilde{\zeta_y}}$ are used to calculate the projection of the image vector $\boldsymbol{\rho_{\varphi}^j}$ on the $y^{th}$ subspace. The projection $\boldsymbol{\widehat{\rho}_y^j}$ of the image vector $\boldsymbol{\rho_{\varphi}^j}$ on the $y^{th}$ subspace is calculated by the following equations:
\begin{equation}
\label{Solv2} 
\boldsymbol{\widehat{\rho}_y^j}=\boldsymbol{\widetilde{\zeta_y}}\boldsymbol{\theta_y^j},\quad \forall\ j\in [N_{P}],\ \forall\ y\in [Y]
\end{equation}
\begin{equation}
\label{Solv3} 
\boldsymbol{\widehat{\rho}_y^j}=\boldsymbol{\widetilde{\zeta_y}}(\boldsymbol{\widetilde{\zeta_y}}'\boldsymbol{\widetilde{\zeta_y}})^{-1}\boldsymbol{\widetilde{\zeta_y}}'\boldsymbol{\rho_{\varphi}^j}
\end{equation}
The projection $\boldsymbol{\widehat{\rho}_y^j}$ can also be termed as the reconstructed image vector for $\boldsymbol{\rho_{\varphi}^j}$.
Equations (\ref{Solv2}) and (\ref{Solv3}) are useful to estimate the image projections in online mode i.e., when all the test data is not available simultaneously. However, for an offline classification scenario, the matrix implementation of the above problem can be efficiently used to decrease the processing time.
Let $\boldsymbol{\Theta_y} \in \mathbb{R}^{N_{\zeta}\times N_{P}}$ be a matrix of parameters. Then the test image set matrix $\boldsymbol{P_{\varphi}}$ can be represented as the combination of $\boldsymbol{\Theta_y}$ and the gallery set matrix $\boldsymbol{\widetilde{\zeta_y}}$:

\begin{equation}
\label{EqtoSolvMat} 
\boldsymbol{P_{\varphi}}=\boldsymbol{\widetilde{\zeta_y}}\boldsymbol{\Theta_y},\quad \forall\ y\in [Y]
\end{equation} 
$\boldsymbol{\Theta_y}$ can be calculated by using the least square estimation using the following equations:
\begin{equation}
\label{Solv4} 
\boldsymbol{\Theta_y} =(\boldsymbol{\widetilde{\zeta_y}}'\boldsymbol{\widetilde{\zeta_y}})^{-1}\boldsymbol{\widetilde{\zeta_y}}'\boldsymbol{P_{\varphi}},\quad \forall\ y\in [Y]
\end{equation}

The projections of the test image vectors on the $y^{th}$ subspace can be calculated as follows:
\begin{equation}
\label{Solv5} 
\boldsymbol{\widehat{P}_y}=\boldsymbol{\widetilde{\zeta_y}}\boldsymbol{\Theta_y},\quad \forall\ y\in [Y]
\end{equation}
\begin{equation}
\label{Solv6} 
\boldsymbol{\widehat{P}_y}=\boldsymbol{\widetilde{\zeta_y}}(\boldsymbol{\widetilde{\zeta_y}}'\boldsymbol{\widetilde{\zeta_y}})^{-1}\boldsymbol{\widetilde{\zeta_y}}'\boldsymbol{P_{\varphi}}
\end{equation} 
where $\boldsymbol{\widehat{P}_y} \in \mathbb{R}^{T\times N_{P}}$ is the  matrix of projections of the image vectors for the test image set $\boldsymbol{P_{\varphi}}$ on the $y^{th}$ subspace. The difference between each test image $\boldsymbol{\rho_{\varphi}^j}$ and its projection $\boldsymbol{\widehat{\rho}_y^j}$, called residual, is calculated using the Euclidean distance:
\begin{equation}
\label{Solv7} 
r_y^j = \left \| \boldsymbol{\rho_{\varphi}^j}-\boldsymbol{\widehat{\rho}_y^j} \right \|_2,\quad \forall\ y\in [Y],\ \forall\ j\in [N_{P}]
\end{equation}  
Once the residuals are calculated we propose three strategies to reach the final decision for the unknown class $\varphi$ of the test set:

\subsection{Majority Voting (MV)}
In majority voting, each image $j$ of the test image set casts an equal weighted vote $\vartheta^j$ for the class $y$, whose regressor $\boldsymbol{\zeta_y}$ results in the minimum value of residual for that test image vector:
\begin{equation}
\label{Eq:MV1} 
\vartheta^j = arg\:\underset{y}{min} (r_y^j)\quad \forall\ j\in [N_p]
\end{equation} 
Then the votes are counted for each class and the class with the maximum number of votes is declared as the unknown class of the test image set.
\begin{equation}
\label{Eq:MV2} 
\varphi_{MV} = mode([\vartheta^1 \vartheta^2 \vartheta^3 ... \vartheta^{N_{P}}])
\end{equation}
In the case of a tie between two or more classes, we calculate the mean of residuals for the competing classes, and the class with the minimum mean of residuals is declared as the output class of the test image set. Suppose classes $1,2$ and $3$ have the same number of votes then:
\begin{equation}
\label{Eq:MV3} 
r_{y(mean)} = \underset{j}{mean} (r_y^j)\quad \forall\ y\in [1,2,3]
\end{equation} 
\begin{equation}
\label{Eq:MV4} 
\varphi_{MV} = arg\:\underset{y}{min} (r_{y(mean)})
\end{equation}
\subsection{Nearest Neighbour Classification (NN)}
In Nearest Neighbour Classification, we first find the minimum residual $r_{y(min)} $ across all classes in $[Y]$ for each of the test image vector.
\begin{equation}
\label{Eq:NN1} 
r_{y(min)} = \underset{j}{min} (r_y^j)\quad \forall\ y\in [Y]
\end{equation} 
Then, the class with the minimum value of $r_{y(min)}$ is declared as the output class of the test image set. 

\begin{equation}
\label{Eq:NN2} 
\varphi_{NN} = arg\:\underset{y}{min} (r_{y(min)})
\end{equation}

 \begin{table*}[htbp]
  \caption{Average percentage classification accuracies for the task of video-based face recognition on UCSD/Honda (Honda)\citep{lee2003video} dataset for different number of gallery and test set images.}
  \centering
    \begin{tabular}{|S|Q|Q|Q|Q|Q|Q|Q|Q|Q|}
    \hline
    \multicolumn{1}{|c|}{\textbf{No. of Images$\rightarrow$}} & \multicolumn{3}{c|}{\textbf{20-20}} & \multicolumn{3}{c|}{\textbf{40-40}} & \multicolumn{3}{c|}{\textbf{All-All}} \\
    \hline
    \multicolumn{1}{|l|}{\textbf{Methods$\downarrow$ \textbackslash $\ $Resolutions$\rightarrow$}} & \textbf{10$\boldsymbol{\times}$10} & \textbf{15$\boldsymbol{\times}$15} & \textbf{20$\boldsymbol{\times}$20} & \textbf{10$\boldsymbol{\times}$10} & \textbf{15$\boldsymbol{\times}$15} & \textbf{20$\boldsymbol{\times}$20} & \textbf{10$\boldsymbol{\times}$10} & \textbf{15$\boldsymbol{\times}$15} & \textbf{20$\boldsymbol{\times}$20} \\
    \hline
    \textbf{AHISD \citep{cevikalp2010face}} & 91.02 & 92.82 & 92.82 & 92.82 & 95.12 & 94.61 & 93.84 & 98.71 & 99.23 \\
    \hline
    \textbf{RNP\citep{yang2013face}} & 94.35 & 95.12 & 95.12 & 95.64 & 97.43 & 96.41 & 95.64 & 99.48 & 99.48 \\
    \hline
    \textbf{SSDML \citep{zhu2013point}} & 80.51 & 83.08 & 83.85 & 84.10 & 85.64 & 85.64 & 89.74 & 90.77 & 91.54 \\
    \hline
    \textbf{DLRC\citep{chen2014dual}} & 92.30  & 92.82  & 92.82  & 90.51  & 93.07  & 92.82  & 94.3  & 96.9  & 97.1  \\
    \hline
    \textbf{DRM \citep{hayat2015deep}} & 31.28 & 26.92 & 28.46 & 81.79    & 83.33 & 79.49 & \textbf{100} & 99.74 & \textbf{100}  \\
    \hline
    \textbf{PLRC \citep{feng2016pairwise}} & 92.82 & 93.07 & 93.84 & 94.35 & 95.12 & 95.38 & 99.48 & \textbf{100} & \textbf{100}\\
    \hline
    \textbf{PDL \citep{pdl}} & 88.97 & 90.77 & 91.28 & 96.41 & 96.41 & 94.87 & 98.72 & 99.74 & 99.49\\
    \hline
    \textbf{DARG \citep{darg}} & 60.00 & 57.95 & 58.72 & 44.36 & 48.72 & 56.15 & 75.90 & 77.95 & 89.74 \\
    \hline
        \textbf{Ours (MV)} & 90.51& 91.80 & 92.05 & 94.36  & 95.13  & 96.15  & \textbf{100} & \textbf{100} & \textbf{100} \\
    \hline
    \textbf{Ours (NN)} & 95.64 & \textbf{96.15} & \textbf{95.90} & 96.41  & \textbf{97.95}  & 97.69  &\textbf{100} & \textbf{100} & \textbf{100}\\
    \hline
    \textbf{Ours (EWV)} & \textbf{95.90} & \textbf{96.15} & \textbf{95.90} & \textbf{97.18} & \textbf{97.95} & \textbf{98.72 }& \textbf{100} & \textbf{100} & \textbf{100} \\
    \hline
    \end{tabular}%
     \vspace{0.2cm}
   
  \label{table:honda}%
\end{table*}%
\subsection{Exponential Weighted Voting (EWV)}
In exponential weighted voting, each image $[N_P]$ in the test image set casts a vote $\omega _y^j$ to each class $y$ in the gallery. We tested with different types of weights including the Euclidean distance, exponential of the Euclidean distance, the inverse of the Euclidean distance and the square of the inverse Euclidean distance. Empirically, it was found that the best performance was achieved when we used the exponential of the Euclidean distance as weights for the votes. Hence, the following equation defines the weight of the vote $\omega _y^j$ of each image $j$ :
\begin{equation}
\label{Eq:EWV1} 
\omega_y^j = e^{-\beta r_y^j},\quad y\in [Y], \ \forall\ j\in [N_{P}]
\end{equation} 
where $\beta$ is a constant. The weights are accumulated for each gallery class $[Y]$ from each image of the test set:
\begin{equation}
\label{Eq:EWV2} 
\varpi_y = \sum_{j=1}^{N_{P}} \omega_y^j,\quad \forall\ y\in [Y]
\end{equation}
The final decision is ruled in the favour of the class $y$ with the maximum accumulated weight from all the images $\boldsymbol{\rho_{\varphi}^j}$ of the test image set $\boldsymbol{P_{\varphi}}$:
\begin{equation}
\label{Eq:EWV3} 
\varphi_{EWV} = arg\:\underset{y}{max} (\varpi_y)
\end{equation}

\subsection{Fast Linear Image Reconstruction} \label{fast}
If the testing is done offline (i.e., all test data is available), then the processing time can substantially be reduced by using Equations (\ref{EqtoSolvMat}), (\ref{Solv4}), (\ref{Solv5}) and (\ref{Solv6}) instead of Equations (\ref{EqtoSolv}), (\ref{Solv1}), (\ref{Solv2}) and (\ref{Solv3}). The computational efficiency of the proposed technique can further be improved by using Moore-Penrose pseudoinverse \citep{penrose1956best,stoer2013introduction} to estimate the inverse of the regressor matrix $\boldsymbol{\widetilde{\zeta_y}}$ at the time of the gallery formation. This step significantly reduces the computations required at test time.

Table \ref{table:testtime} shows that many fold gain in computational efficiency was achieved by the fast linear image reconstruction technique for the YouTube Celebrity dataset \citep{kim2008face}. Let $\boldsymbol{\widetilde{\zeta_{y}^{inv}}}$ be the pseudoinverse of the regressor $\boldsymbol{\widetilde{\zeta_y}}$ calculated at the time of gallery formation. Equation (\ref{EqtoSolvMat}) can then be solved at the test time as:

\begin{equation}
\label{pseudoinverse}
\boldsymbol{\Theta_y}=\boldsymbol{\widetilde{\zeta_{y}^{inv}}} \boldsymbol{P_{\varphi}}
\end{equation}

\begin{equation}
\label{SolvPinv} 
\boldsymbol{\widehat{P}_y}=\boldsymbol{\widetilde{\zeta_y}}(\boldsymbol{\widetilde{\zeta_{y}^{inv}}} \boldsymbol{P_{\varphi}})
\end{equation} 

\section{Experiments and Analysis}\label{experiments}
We rigorously evaluated the proposed technique for the tasks of surveillance, video-based face recognition and object recognition.
We evaluated our technique on three surveillance databases, namely SCFace \citep{scfacegrgic2011}, FR\_SURV \citep{frsurv} and Choke Point \citep{choke} datasets.
For video-based face recognition, we used the publicly available and challenging databases, namely Honda/UCSD Dataset \citep{lee2003video}, CMU Motion of Body Dataset (CMU MoBo) \citep{gross2001cmu} and Youtube Celebrity Dataset (YTC) \citep{kim2008face}. Washington RGBD Object \citep{wrgbd} and ETH-80 \citep{leibe2003analyzing} datasets were used for the task of object recognition. We also performed experiments for limited number of images in the gallery and the test datasets to verify the performance of the proposed technique for the conditions where only scarce amount of data is available. All the experiments were performed at the resolutions of $20\times 20$ pixels, $15\times 15$ pixels and $10\times 10$ pixels to establish the performance of our technique at extremely low resolution.

We compared our technique with several prominent image set classification methods. These techniques include the Affine Hull based Image Set Distance (AHISD) \citep{cevikalp2010face}, Regularized Nearest Points (RNP) \citep{yang2013face}, Set to Set Distance Metric Learning (SSDML) \citep{zhu2013point}, Dual Linear Regression Classifier (DLRC) \citep{chen2014dual}, Deep Reconstruction Models (DRM) \citep{hayat2015deep}, Pairwise Linear Regression Classifier (PLRC) \citep{feng2016pairwise}, Prototype Discriminative Learning (PDL) \citep{pdl} and Discriminant Analysis on Riemannian Manifold of Gaussian Distributions (DARG) \citep{darg}. We followed the standard experimental protocols. For the compared methods, we used the implementations provided by the respective authors. For all of the compared methods, the parameter settings which were recommended in the respective papers were used. For DARG \citep{darg}, we kept 95\% energy of the PCA features. Since DLRC \citep{chen2014dual} and PLRC \citep{feng2016pairwise} can work with only small datasets, we selected the number of training and testing images for these techniques depending on the best accuracy in each of the datasets.

\subsection{Surveillance}
\subsubsection{SCFace Dataset}
\begin{figure*}[t]
\begin{center}
   \includegraphics[width=0.8\linewidth]{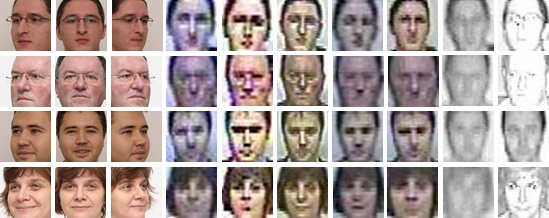}
\end{center}
   \caption{Faces detected by the Viola and Jones face detection algorithm of four individuals from the challenging SCFace Dataset \citep{scfacegrgic2011}. The first three columns show the Gallery images while the remaining columns show the test images. The large difference between the quality of the gallery and test images is evident.}
\label{fig:scface}
\end{figure*}

The SCface dataset \citep{scfacegrgic2011} contains static images of human faces. The main purpose of SCface dataset is to test face recognition algorithms in real world scenarios. This dataset presents the challenges of different training and testing environments, uncontrolled illumination, low resolution, less gallery and test data, head pose orientation and a large number of classes.
Five commercially available video surveillance cameras of different qualities and various resolutions were used to capture the images of 130 individuals in uncontrolled indoor environments. Different quality cameras present a real world surveillance scenario for face recognition. Due to the large number of individuals, there is a very low probability of results obtained by pure chance. 
Images were taken under uncontrolled illumination conditions with outdoor light coming from a window as the only source of illumination. Two of the surveillance cameras record both visible spectrum and IR night vision images, which are also included in the dataset.
Images were captured from various distances for each individual.
Another challenge in the dataset is the head pose i.e., the camera is above the subject's head to mimic a commercial surveillance systems while for the gallery images the camera is in front of the individual. 
For the gallery set, individuals were photographed with a high quality digital photographer's camera at close range in controlled conditions (standard indoor lighting, adequate use of flash to avoid shades, high resolution of images) to mimic the conditions used by law enforcement agencies. Different views were captured for each individual from -90 degrees to +90 degrees. Figure \ref{fig:scface} clearly presents the difference in the quality of images in the gallery images and test images of four individuals in the SCFace dataset.
 
We used the images taken in controlled conditions for the gallery sets while all other images were used to form the test image sets. The faces in the images were detected using the Viola and Jones face detection algorithm \citep{viola2004robust}. The gallery sets have a maximum of six images. The images were converted to grayscale and resized to $10\times10$, $15\times15$ and $20\times20$. Table \ref{table:surveillance} shows the classification accuracies of our technique compared to the other state of the art techniques. Our technique achieved the best classification accuracy on all resolutions. This can be attributed to the fact that there is not sufficient data available for training, so the techniques which depend heavily on data availability e.g., \citep{hayat2015deep} fail to produce good results.

\subsubsection{FR\_SURV Dataset}
\begin{figure*}[t]
\begin{center}
   \includegraphics[width=0.8\linewidth]{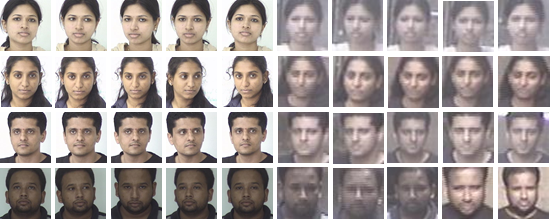}
\end{center}
   \caption{Random images of four individuals from the challenging FR\_Surv Dataset \citep{frsurv}. The first five columns show the gallery images while the remaining columns show the test images. The difference in the quality of the gallery and the test images can clearly be observed.}
\label{fig:frsurve}
\end{figure*}

The FR\_SURV dataset \citep{frsurv} contains images of 51 individuals. The main purpose of this dataset is to test face recognition algorithms in long distance surveillance scenarios. This dataset presents the challenges of uncontrolled illumination, low resolution, blur, noise, less gallery and test data, head pose orientation and a large number of classes. Unlike most other datasets, the gallery and test images were captured in different conditions. The gallery set contains twenty high resolution images of each individual captured from a close distance in controlled indoor conditions. The images contain slight head movements and the camera is placed parallel to the head position. There is a significant difference between the resolutions of the gallery and the test images. The test images were captured in outdoor conditions from a distance of 50m - 100m. The camera was placed at the height of around 20m - 25m. Due to the use of commercial surveillance camera and the large distance between the camera and the individual, the test images are of very low quality which makes it difficult to identify the individual. The difference in the gallery and test images of the FR\_SURV dataset in Figure \ref{fig:frsurve} shows the challenges of the dataset.

We used the indoor images for the gallery sets while the outdoor images were used to form the test image sets. Viola and Jones face detection algorithm \citep{viola2004robust} was used to detect faces in the images. The images were converted to grayscale and resized to $10\times10$, $15\times15$ and $20\times20$. The classification accuracies of our technique compared to the other image set classification techniques are shown in Table \ref{table:surveillance}. Although the results of all techniques are low on this dataset, our technique achieved the best performance on all resolutions. The percentage gap in the performance of our technique and the second best technique is more than 15\%.  The techniques which require a large training data fail to produce any considerable results on this dataset.

\begin{table*}[htbp]
\caption{Average percentage classification accuracies for the task of video-based face recognition on CMU MoBo (MoBo)\citep{gross2001cmu} dataset for different sizes of gallery and test images.}
  \centering
    \begin{tabular}{|c|Q|Q|Q|Q|Q|Q|Q|Q|Q|}
    \hline
    \multicolumn{1}{|c|}{\textbf{No. of Images$\rightarrow$}} & \multicolumn{3}{c|}{\textbf{20-20}} & \multicolumn{3}{c|}{\textbf{All-All}} \\
    \hline
    \multicolumn{1}{|l|}{\textbf{Methods$\downarrow$ \textbackslash $\ $Resolutions$\rightarrow$}} & \textbf{10$\boldsymbol{\times}$10} & \textbf{15$\boldsymbol{\times}$15} & \textbf{20$\boldsymbol{\times}$20} & \textbf{10$\boldsymbol{\times}$10} & \textbf{15$\boldsymbol{\times}$15} & \textbf{20$\boldsymbol{\times}$20} \\
    \hline
    \textbf{AHISD \citep{cevikalp2010face}} & 62.77 & 71.66 & 76.80 & 93.89 & 96.25 & 96.81\\
    \hline
    \textbf{RNP\citep{yang2013face}} & 66.25 & 72.5 & 77.63 & 92.36 & 95.69 & 97.36 \\
    \hline
    \textbf{SSDML \citep{zhu2013point}} & 73.47 & 73.06 & 73.47 & 96.39 & 96.67 & 96.25\\
    \hline
    \textbf{DLRC\citep{chen2014dual}} & 80.69  & 82.08  & 80.00  & 96.2  & 97.2  & 97.7\\
    \hline
    \textbf{DRM \citep{hayat2015deep}} & 26.94 & 40.14 & 38.06 & 90    & 90.56 & 91.81\\
    \hline
    \textbf{PLRC \citep{feng2016pairwise}} & 79.72 & 80.41 & 79.16 & 96.38 & 96.38 & 96.94\\
    \hline
    \textbf{PDL \citep{pdl}} & 70.28 & 72.50 & 74.17 & 98.33 & 98.33 & 98.47\\
    \hline
    \textbf{DARG \citep{darg}} & 37.92 & 39.58 & 40.42 & 71.81 & 72.92 & 76.53\\
    \hline
        \textbf{Ours (MV)} & 73.06 & 74.31 & 74.17 & 90.97  & 90.70  & 91.53\\
    \hline
    \textbf{Ours (NN)} & \textbf{80.97} & \textbf{83.06} & \textbf{82.92} & \textbf{98.61}  & \textbf{99.31}  & \textbf{99.03} \\
    \hline
    \textbf{Ours (EWV)} &80.56 & \textbf{83.06} & \textbf{82.92} & \textbf{98.61} & \textbf{99.31} & \textbf{99.03}\\
    \hline
    \end{tabular}%
     \vspace{0.2cm}
    
  \label{table:mobo}%
\end{table*}%
\subsubsection{Choke Point Dataset}

The Choke Point dataset \citep{choke} consists of videos of 29 individuals captured using  surveillance cameras. Three cameras at different position simultaneously record the entry or exit of a person from 3 viewpoints. The main purpose of this dataset was to identify or verify persons under real-world surveillance conditions. The cameras were placed above the doors to capture individuals walking through the door in a natural manner. This dataset presents the challenges of variations in illumination, background, pose and sharpness. The 48 video sequences of the dataset contain 64,204 face images at a frame rate of 30 fps with an image resolution of $800 \times 600$ pixels. In all sequences, only one individual is present in the image at a time. The sequences contain different backgrounds depending on the door and whether the individual is entering or leaving the room. Overall, 25 individuals appear in 16 video sequences while 4 appear in 8 video sequences.

We used the Viola and Jones face detection algorithm \citep{viola2004robust} to detect faces in the images. The images were converted to grayscale and resized to $10\times10$, $15\times15$ and $20\times20$. As recommended by the authors of the dataset for the evaluation protocol \citep{choke}, we used only images from the camera which captured the most frontal faces for each video sequence. 
This results in 16 image sets for 25 individuals and 8 image sets for 4 individuals. We randomly selected two image sets of each individual for the gallery set while the rest of the image sets were used as test image sets. The experiment was repeated ten times with a random selection of gallery and test image sets to improve the generalization of results. We achieved average classification accuracies of $71.23\%$, $73.85\%$ and $72.75\%$ for $10\times10$, $15\times15$ and $20\times20$ resolutions, respectively. In order to test the performance of image set classification methods under the challenges of limited gallery and test data, we reduced the size of image sets to 20 by retaining the first 20 images of each image set. Table \ref{table:surveillance} shows the classification accuracies on the Choke Point Dataset. Our technique achieved the best classification accuracies compared to the other image set techniques.

\subsection{Video-based Face Recognition}
For video-based face recognition, we used the most popular image set classification datasets. For this section we also carried out extensive experiments using the full length image sets, in addition to the experiments which use limited gallery and test data, in order to demonstrate the suitability of our technique for both scenarios. In experiments using full length image sets, we reduced the gallery data, for our technique, whenever the number of images in the gallery set was larger than the number of pixels in the downsampled images. This ensures that the number of features is always larger than the number of images (see Section \ref{technique} for details) for our technique. 
\subsubsection{UCSD/Honda Dataset}
The UCSD/Honda Dataset \citep{lee2003video} was developed to evaluate the face tracking and recognition algorithms. The dataset  consists of 59 videos of 20 individuals. The resolution of each video sequence is $640\times 480$. Of the 20 individuals, all but one have at least two videos in the dataset. All videos contain significant head rotations and pose variations. Moreover, some of the video sequences also contain partial occlusions in some frames. We followed the standard experimental protocol used in \citep{lee2003video,wang2008manifold,wang2009manifold,hu2012face,hayat2015deep}. The face from each frame of videos was automatically detected using the Viola and Jones face detection algorithm \citep{viola2004robust}. The detected faces were down sampled to the resolutions of $10\times 10$, $15\times 15$ and $20\times 20$ and converted to grayscale. Histogram equalization was applied to increase the contrast of images. For the gallery image sets, we randomly selected one video of each individual. The remaining 39 videos were used as the test image sets. For our technique, we randomly selected a small number of frames to keep the number of gallery images considerably smaller than the number of pixels (refer to Section \ref{technique}). To improve the consistency in scores we repeated the experiment 10 times with a different random selection of gallery images, gallery image sets and test image sets. Our technique achieved $ 100\% $ classification accuracy on all the resolutions while using a significantly less number of gallery images. We also carried out experiments for less gallery and test data by retaining the first $20$ and first $40$ images of each image set. Our technique achieved the best classification performance in all of the testing protocols. Table \ref{table:honda} summarizes the average identification rates of our technique compared to other image set classification techniques on the Honda dataset.

\subsubsection{CMU MoBo Dataset}
\begin{figure*}[t]
\begin{center}
   \includegraphics[width=0.8\linewidth]{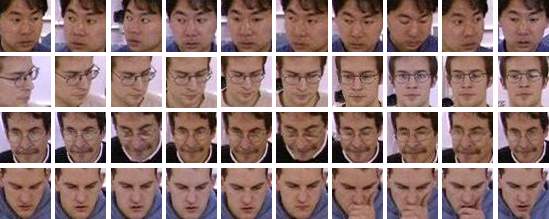}
\end{center}
   \caption{Face images detected by the Viola and Jones face detection Algorithm of four individuals from CMU MoBo dataset \citep{gross2001cmu}.}
\label{fig:mobo}
\end{figure*}

The original purpose of the CMU Motion of Body Dataset (CMU MoBo) was to advance biometric research on human gait analysis \citep{gross2001cmu}. It contains videos of 25 individuals walking on a treadmill, captured from six different viewpoints. Except the last one, all subjects in the dataset have four videos following different walking patterns. The four walking patterns are slow walk, fast walk, inclined walk and holding a ball while walking. Figure \ref{fig:mobo} shows random images of four individuals in the CMU MoBo dataset detected by the Viola and Jones face detection algorithm \citep{viola2004robust}

Following the same procedure as previous works \citep{chen2014dual,hayat2015deep,feng2016pairwise}, we used the videos of the 24 individuals which contain all the four walking patterns. Only the videos from the camera in front of the individual were used for image set classification. The frames of each video were considered as an image set. We followed the standard protocol \citep{wang2008manifold,cevikalp2010face,hu2012face,hayat2015deep}, and randomly selected the video of one walking pattern of each individual for the gallery image set. The remaining videos were considered as test image sets. As mentioned in Section \ref{technique}, the number of images should be less than or equal to the number of pixels in the downsampled images. In practice, the number of images should be considerably less than the number of pixels. For our technique, we randomly selected a small number of frames from each gallery video to form the gallery image sets. The face from each frame was automatically detected using the Viola and Jones face detection algorithm \citep{viola2004robust}. The images were resampled to the resolutions of $10\times 10$, $15\times 15$ and $20\times 20$ and converted to grayscale. Histogram equalization was applied to increase the contrast of images. 
Unlike most other techniques, we used raw images for our technique and did not extract any features, such as LBP features used in \citep{cevikalp2010face,chen2014dual,hayat2015deep,feng2016pairwise}. The experiments were repeated 10 times with different random selections of the gallery and the test sets. We also used different random selections of the gallery images in each round to make our testing environment more challenging. Our technique achieved the best performance on this dataset, with greater than 99\% average classification accuracy. Then we reduced the number of images in each image set to the first 20. Our technique achieved the best classification accuracy in both experimental protocols. Table \ref{table:mobo} provides the average accuracy of our technique along with a comparison with other methods.
\begin{table*}[htbp]
\caption{Average percentage classification accuracies for the task of video-based face recognition on YouTube Celebrity (YTC) \citep{kim2008face} dataset for different sizes of gallery and test image sets.}
  \centering
    \begin{tabular}{|c|Q|Q|Q|Q|Q|Q|Q|Q|Q|}
    \hline
    \multicolumn{1}{|c|}{\textbf{No. of images$\rightarrow$}} & \multicolumn{3}{c|}{\textbf{60-20}} & \multicolumn{3}{c|}{\textbf{All-All}} \\
    \hline
    \multicolumn{1}{|l|}{\textbf{Methods$\downarrow$ \textbackslash $\ $Resolutions$\rightarrow$}} & \textbf{10$\boldsymbol{\times}$10} & \textbf{15$\boldsymbol{\times}$15} & \textbf{20$\boldsymbol{\times}$20} & \textbf{10$\boldsymbol{\times}$10} & \textbf{15$\boldsymbol{\times}$15} & \textbf{20$\boldsymbol{\times}$20} \\
    \hline
    \textbf{AHISD \citep{cevikalp2010face}} & 59.71 & 59.36 & 58.58 & 43.9  & 55.88 & 54.18 \\
    \hline
    \textbf{RNP\citep{yang2013face}} & 58.22 & 57.65 & 56.59 & 65.46 & 65.74 & 65.46 \\
    \hline
    \textbf{SSDML \citep{zhu2013point}} & 60.5 & 61.42 & 60.14 & 63.26 & 65.25 & 64.4 \\
    \hline
    \textbf{DLRC\citep{chen2014dual}} & 57.02  & 61.13  & 60.42  & 60.14 & 64.4  & 63.9 \\
    \hline
    \textbf{DRM \citep{hayat2015deep}} & 39.93    & 46.52 & 51.91 & 49.36 & 57.66 & 61.06 \\
    \hline
    \textbf{PLRC \citep{feng2016pairwise}} & 60.19 & 59.27 & 59.48 & 62.08 & 62.51 & 62.59 \\
    \hline
    \textbf{PDL \citep{pdl}} & 53.05 & 52.77 & 53.40 & 54.61 & 54.47 & 54.82 \\
    \hline
    \textbf{DARG \citep{darg}} & 39.11 & 41.74 & 43.39 & 50.02 & 53.02 & 51.74 \\
    \hline

    \textbf{Ours (MV)} & 57.38 & 58.23  & 56.95  & 60.71 & 60.64  & 59.29 \\
    \hline
    \textbf{Ours (NN)} & \textbf{60.78}  & \textbf{61.49}  & \textbf{61.35}  & \textbf{66.45} & \textbf{66.81}  & \textbf{66.46} \\
    \hline
    \textbf{Ours (EWV)} & 59.65 & 60.71 & 60.78 & 65.25 & 66.24 & 65.89 \\
    \hline
    \end{tabular}%
     \vspace{0.2cm}
    
  \label{table:ytc}%
\end{table*}%
\subsubsection{Youtube Celebrity Dataset}
The YoutubeCelebrity (YTC) Dataset \citep{kim2008face} contains cropped clips from videos of 47 celebrities and politicians. In total, the dataset contains 1910 video clips. The videos are downloaded from the YouTube and are of very low quality. Due to their high compression rates, the video clips are very noisy and have a low resolution. This makes it a very challenging dataset.
The Viola and Jones face detection algorithm \citep{viola2004robust} failed to detect any face in a large number of frames. Therefore, the Incremental Learning Tracker \citep{ross2008incremental} was used to track the faces in the video clips. 
To initialize the tracker, we used the initialization parameters\footnote{\url{http://seqam.rutgers.edu/site/index.php?option=com_content&view=article&id=64&Itemid=80}} provided by the authors of \citep{kim2008face}. This approach, however, produced many tracking errors due to the low image resolution and noise. Although the cropped face region was not uniform across frames, we decided to use the automatically tracked faces without any refinement to make our experimental settings more challenging. 

\begin{figure*}[t]
\begin{center}
   \includegraphics[width=0.8\linewidth]{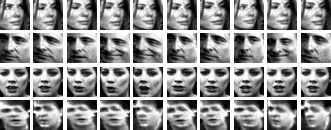}
\end{center}
   \caption{Random histogram equalized, grayscale images of four celebrities from YouTube Celebrity dataset \citep{kim2008face}.}
\label{fig:ytc}
\end{figure*}

We followed the standard protocol which was also followed in \citep{wang2008manifold,wang2009manifold,wang2012covariance,hu2012face,hayat2015deep}, for our experiments. The dataset was divided into five folds with a minimal overlap between the various folds. Each fold consists of 9 video clips per individual, to make a total of 423 video clips. Three videos of each individual were randomly selected to form the gallery set while the remaining six were used as six separate test image sets. All the tracked face images were resampled to the resolutions of $10\times 10$, $15\times 15$ and $20\times 20$ and converted to grayscale. We applied histogram equalization to enhance the image contrast. Figure \ref{fig:ytc} shows histogram equalized images of four celebrities in the YouTube Celebrity dataset. For the gallery formation in our technique, we randomly selected a small number of images from each of the three gallery videos per individual per fold in order to keep the number of gallery images less than the number of pixels (features). Our technique achieved the highest classification accuracy among all the methods, while using significantly less gallery data compared to the other methods. 
We also carried out experiments for less gallery and test data by selecting the first 20 images from each image set. If any video clip had less than 20 frames, all the frames of that video clip were used for gallery formation. In this way each gallery set had a maximum of 60 images while each test set had a maximum of 20 images. Our technique achieved the best classification accuracies for all resolutions.

Table \ref{table:ytc} summarizes the average accuracies of the different techniques on the YouTube Celebrity dataset.

\begin{table*}[htbp]
\caption{Average percentage classification accuracies for the task of object recognition on Washington RGB-D Object \citep{wrgbd} and ETH-80 \citep{leibe2003analyzing} datasets.}
  \centering
     \begin{tabular}{|S|V|V|V|V|V|V|}
    \hline
    \multicolumn{1}{|c|}{\textbf{Datasets$\rightarrow$}} & \multicolumn{3}{c|}{\textbf{Washington RGBD Object}} & \multicolumn{3}{c|}{\textbf{ETH-80}} \\
    \hline
    \multicolumn{1}{|l|}{\textbf{Methods$\downarrow$ \textbackslash $\ $Resolutions$\rightarrow$}} & \textbf{10$\boldsymbol{\times}$10} & \textbf{15$\boldsymbol{\times}$15} & \textbf{20$\boldsymbol{\times}$20} & \textbf{10$\boldsymbol{\times}$10} & \textbf{15$\boldsymbol{\times}$15} & \textbf{20$\boldsymbol{\times}$20} \\
    \hline
    \textbf{AHISD \citep{cevikalp2010face}} & 53.78 & 57.32 & 57.37 & 73.25 & 85.25 & 87.5 \\
    \hline
    \textbf{RNP\citep{yang2013face}} & 52.72  & 55.60  & 57.17 & 74    & 74.25 & 79 \\
    \hline
    \textbf{SSDML \citep{zhu2013point}} & 44.39 & 46.31 & 43.33 & 73.25 & 69    & 67.75 \\
    \hline
    \textbf{DLRC\citep{chen2014dual}} & 50.91  & 56.76  & 57.77  & 76.2  & 85.2  & 87.2 \\
    \hline
    \textbf{DRM \citep{hayat2015deep}} & 34.85 & 51.92  & 61.72 & 88.75 & 90.25 & 88.75 \\
    \hline
    \textbf{PLRC \citep{feng2016pairwise}} & 55.61 & 58.13  & 59.17 & 80.93 & 86.47 & 87.44 \\
    \hline
    \textbf{PDL \citep{pdl}} & 43.23 & 48.03 & 46.92 & 76.50 & 82.75 & 83.50 \\
    \hline
    \textbf{DARG \citep{darg}} & 21.52 & 22.22 & 11.67 & 57.00 & 54.50 & 51.25 \\
    \hline
    \textbf{Ours (MV)} & 61.06  & 63.74  & 62.58  & 89.75  & 94.75  & \textbf{96} \\
    \hline
    \textbf{Ours (NN)} & 53.13  & 56.31  & 53.94  & 74.25  & 81  & 81 \\
    \hline
    \textbf{Ours (EWV)} & \textbf{61.52} & \textbf{64.80} & \textbf{63.33} & \textbf{90.5} & \textbf{95.25} & 95.75 \\
    \hline
    \end{tabular}%
     \vspace{0.2cm}
    
  \label{table:object}%
\end{table*}%

\subsection{Object Recognition}
\subsubsection{Washington RGBD Object Dataset}
The Washington RGB-D Object Dataset \citep{wrgbd} is a large dataset, which consists of common household objects. The dataset contains 300 objects in 51 categories. The dataset was recorded at a resolution of $640 \times 480$ pixels. The objects were placed on a turntable and video sequences were captured at 30 fps for a complete rotation. A camera mounted at 3 different heights was used to capture 3 video sequences for each object so that the object is viewed from different angles.
 
We used the RGB images of the dataset in our experiments. Using the leave one out protocol defined by the authors of the dataset for object class recognition \citep{wrgbd}, we achieved average classification accuracies of $83.33\%$, $83.73\%$ and $84.51\%$ for $10\times10$, $15\times15$ and $20\times20$ resolutions respectively. To evaluate the proposed technique for less gallery and test data, 20 images were randomly selected from the three video sequences of each instance to form a single image set. Two image sets of each class were randomly selected to form the gallery set, while the remaining sets were used as test sets. The images were converted to grayscale and resized to $10\times10$, $15\times15$ and $20\times20$ resolutions.  Table \ref{table:object} shows the classification accuracies of the image set classification techniques on this dataset. Our technique achieved the best classification accuracies at all the resolutions.

\subsubsection{ETH-80 Dataset}
\begin{figure*}[t]
\begin{center}
   \includegraphics[width=0.8\linewidth]{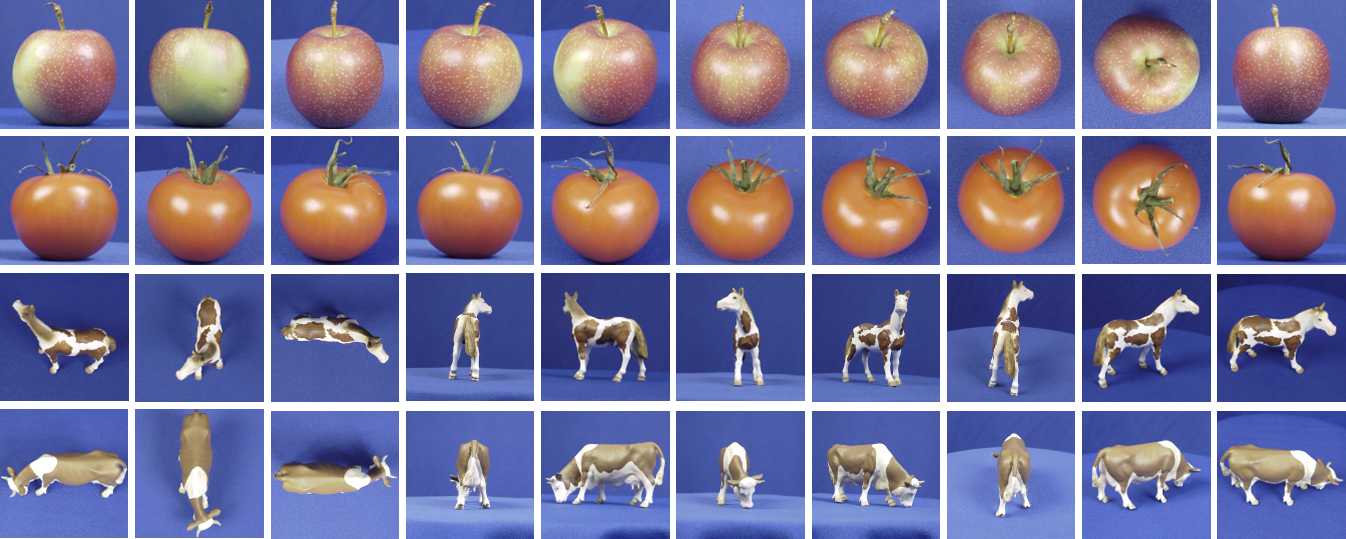}
\end{center}
   \caption{Random images of four of the classes from ETH-80 dataset. The resemblance between the images of apple (first row) and tomatoe (scond row), and between the images of horse (third row) and cow (fourth row) shows the challenge of inter-class similarity \citep{leibe2003analyzing}.}
\label{fig:eth}
\end{figure*}

The ETH-80 dataset \citep{leibe2003analyzing}
contains eight object categories: apples, pears, tomatoes, cups, cars, horses, cows and dogs. There are ten different image sets in each object category. Each image set contains images of an object taken from 41 different view angles in 3D.
We used the cropped images, which contain the object without any border area, provided by the dataset. Figure \ref{fig:eth} shows few images of four of the classes from ETH-80 dataset. 

The images were resized to the resolutions of $10\times10$, $15\times15$ and $20\times20$. We used grayscale raw images for our technique. Similar to \citep{kim2007discriminative,wang2009manifold,wang2012covariance,hayat2015deep}, five image sets of each object category were randomly selected to form the gallery set while the other five are considered to be independent test image sets. In order to improve the generalization of the results, the experiments were repeated 10 times for different random selections of gallery and test sets. Table \ref{table:object} summarizes the results of our technique compared to other methods. We achieved superior classification accuracy compared to all the other techniques at all resolutions. 

\section{Computational Time Analysis}\label{timing_analysis}
The fast approach of the proposed technique requires the least computational time, compared to other image set techniques. In contrast to the significant training times of SSDML \citep{zhu2013point}, DRM \citep{hayat2015deep} and DARG \citep{darg}, the proposed technique does not require any training (Table \ref{table:traintime}). Hence, our technique is one of the fastest to set-up a face recognition system. Our technique is also the fastest at test time, especially for limited gallery and test data. The only method with a comparable efficiency at test time is PDL \citep{pdl}. Table \ref{table:testtime} shows the testing times that are required per fold for the YouTube Celebrity dataset for various methods using a Core i7 3.40GHz CPU with 24 GB RAM. Although our technique reconstructs each image of the test image set from all the gallery image sets, due to its efficient matrix representation (refer to Section \ref{technique} and Section \ref{fast}), we achieved a timing efficiency that is superior to the other methods. Moreover, due to the inherent nature of our technique it is highly parallelizable. It can be effectively deployed on embedded systems. The timing efficiency of the proposed technique can further be improved through parallel processing or GPU computation.

\begin{table*}[htbp]
\caption{Training time in seconds per fold required by different methods for the task of video-based face recognition on YouTube Celebrity (YTC) \citep{kim2008face} dataset for different sizes of training and test image sets. {\footnotesize NR shows that the method does not require training. {\large *} Includes time required to calculate LBP features.}}
  \centering
    \begin{tabular}{|c|Q|Q|Q|Q|Q|Q|Q|Q|Q|}
    \hline
    \multicolumn{1}{|c|}{\textbf{No. of images$\rightarrow$}} & \multicolumn{3}{c|}{\textbf{60-20}} & \multicolumn{3}{c|}{\textbf{All-All}} \\
    \hline
    \multicolumn{1}{|l|}{\textbf{Methods$\downarrow$ \textbackslash $\ $Resolutions$\rightarrow$}} & \textbf{10$\boldsymbol{\times}$10} & \textbf{15$\boldsymbol{\times}$15} & \textbf{20$\boldsymbol{\times}$20} & \textbf{10$\boldsymbol{\times}$10} & \textbf{15$\boldsymbol{\times}$15} & \textbf{20$\boldsymbol{\times}$20} \\
    \hline
    \textbf{AHISD \citep{cevikalp2010face}} & NR & NR  & NR  & NR & NR  & NR \\
    \hline
    \textbf{RNP\citep{yang2013face}} & NR & NR  & NR  & NR & NR  & NR \\
    \hline
    \textbf{SSDML \citep{zhu2013point}} & 9.74 & 18.89 & 21.61 & 101.69 & 193.37 & 278.63 \\
    \hline
    \textbf{DLRC\citep{chen2014dual}} & NR  & NR  & NR  & NR & NR  & NR \\
    \hline
    \textbf{DRM* \citep{hayat2015deep}} & 2194.28 & 2194.36 & 2194.40 & 4149.87 & 4157.23 & 4164.63  \\
    \hline
    \textbf{PLRC \citep{feng2016pairwise}} & NR & NR  & NR  & NR & NR  & NR  \\
    \hline
    \textbf{PDL \citep{pdl}} & 5.10 & 5.18 & 5.23 & 60.33 & 60.71 & 62.18  \\
    \hline
    \textbf{DARG \citep{darg}} & 7.24 & 7.40 & 7.65 & 80.45 & 109.08 & 165.22  \\
    \hline
    \textbf{Ours } & NR & NR  & NR  & NR & NR  & NR \\
    \hline
        \textbf{Ours (Fast)} & NR & NR & NR & NR & NR & NR \\
    \hline
    \end{tabular}%
     \vspace{0.2cm}
    
  \label{table:traintime}%
\end{table*}%
   
   \begin{table*}[htbp]
   \caption{Testing time in seconds per fold required by different methods for the task of video-based face recognition on YouTube Celebrity (YTC) \citep{kim2008face} dataset for different sizes of training and test image sets. Top-2 testing time for each setting are shown in bold. {\large *} Includes time required to calculate LBP features.}
  \centering
    \begin{tabular}{|c|Q|Q|Q|Q|Q|Q|Q|Q|Q|}
    \hline
    \multicolumn{1}{|c|}{\textbf{No. of images$\rightarrow$}} & \multicolumn{3}{c|}{\textbf{60-20}} & \multicolumn{3}{c|}{\textbf{All-All}} \\
    \hline
    \multicolumn{1}{|l|}{\textbf{Methods$\downarrow$ \textbackslash $\ $Resolutions$\rightarrow$}} & \textbf{10$\boldsymbol{\times}$10} & \textbf{15$\boldsymbol{\times}$15} & \textbf{20$\boldsymbol{\times}$20} & \textbf{10$\boldsymbol{\times}$10} & \textbf{15$\boldsymbol{\times}$15} & \textbf{20$\boldsymbol{\times}$20} \\
    \hline
    \textbf{AHISD \citep{cevikalp2010face}} & 22.78 & 30.09 & 33.01 & 63.39 & 233.37 & 568.34 \\
    \hline
    \textbf{RNP\citep{yang2013face}} & 8.16 & 14.20 & 16.07 & 21.35 & 31.83 & 43.65 \\
    \hline
    \textbf{SSDML \citep{zhu2013point}} & 17.26 & 31.75 & 32.06 & 183.36 & 371.79 & 498.54 \\
    \hline
    \textbf{DLRC\citep{chen2014dual}} & 6.6262  & 12.40  & 23.57  & 303.55 & 347.55  & 431.23 \\
    \hline
    \textbf{DRM* \citep{hayat2015deep}} & 89.15 & 89.46 & 89.69 & 381.82 & 386.65 & 391.97 \\
    \hline
    \textbf{PLRC \citep{feng2016pairwise}} & 27.91 & 61.92 & 104.62 & 203.44 & 248.00 & 299.58 \\
    \hline
    \textbf{PDL \citep{pdl}} & \textbf{2.48} & \textbf{2.84} & \textbf{2.90} & \textbf{5.61} & \textbf{6.46} & \textbf{6.51} \\
    \hline
    \textbf{DARG \citep{darg}} & 9.30 & 11.67 & 12.25 & 110.87 & 255.11 & 478.33 \\
    \hline
    \textbf{Ours} & 4.75 & 9.71  & 14.27  & 6.55 & 13.60 & 25.37 \\
    \hline
    \textbf{Ours (Fast)} & \textbf{0.52} & \textbf{1.28} & \textbf{1.89} & \textbf{2.22} & \textbf{4.61} & \textbf{9.06} \\
    \hline
    \end{tabular}%
     \vspace{0.2cm}
    
  \label{table:testtime}%
\end{table*}%

\section{Conclusion and Future Direction}\label{conclusion}
In this paper, we have proposed a novel image set classification technique, which exploits linear regression for efficient image set classification. The images from the test image set are projected on linear subspaces defined by down sampled gallery image sets. The class of the test image set is decided using the residuals from the projected test images.
Extensive experimental evaluations for the tasks of surveillance, video-based face and object recognition, demonstrate the superior performance of our technique under the challenges of low resolution, noise and less gallery and test data, commonly encountered in surveillance settings. Unlike other techniques, the proposed technique has a consistent performance for the different tasks of surveillance, video-based face recognition and object recognition. The technique requires less gallery data compared to other image set classification methods and works effectively at very low resolution.

Our proposed technique can easily be scaled from processing one frame at a time (for live video acquisition) to the processing of all the test data at once (for a faster performance). It can also be quickly and easily deployed as it is very easy to add new classes in our technique (by just arranging images in a matrix) and it does not require any training or feature extraction. Also in the case of resource rich scenarios, our technique can easily be parallelized to further increase the timing efficiency. These advantages make our technique ideal for practical applications such as surveillance. For future research, we plan to integrate non-linear regression and automatic image selection to make our approach more robust to outliers and to further improve its classification accuracy.

\section*{acknowledgements}
This work is partially funded by SIRF Scholarship from the University of Western Australia (UWA) and Australian Research Council (ARC) grant DP150100294 and DP150104251.

Portions of the research in this paper use the SCface database \citep{scfacegrgic2011} of facial images. Credit is hereby given to the University of Zagreb, Faculty of Electrical Engineering and Computing for providing the database of facial images.

Portions of the research in this paper use the Choke Point dataset \citep{choke} which is sponsored by NICTA. NICTA is funded by the Australian Government as represented by the Department of Broadband, Communications and the Digital Economy, as well as the Australian Research Council through the ICT Centre of Excellence program.

\balance
\bibliographystyle{spbasic}
\bibliography{bibForJournal}
\end{document}